# A Nepali Rule Based Stemmer and its performance on different NLP applications

Pravesh Koirala
*Department of Electronics and Computer Engineering*
*Institute Of Engineering, Pulchowk*
Kathmandu, Nepal
praveshkoirala@gmail.com

Aman Shakya
*Department of Electronics and Computer Engineering*
*Institute Of Engineering, Pulchowk*
Kathmandu, Nepal
aman.shakya@ioe.edu.np

*Abstract*—Stemming is an integral part of Natural Language Processing (NLP). It's a preprocessing step in almost every NLP application. Arguably, the most important usage of stemming is in Information Retrieval (IR). While there are lots of work done on stemming in languages like English, Nepali stemming has only a few works. This study focuses on creating a Rule Based stemmer for Nepali text. Specifically, it is an affix stripping system that identifies two different class of suffixes in Nepali grammar and strips them separately. Only a single negativity prefix न is identified and stripped. This study focuses on a number of techniques like exception word identification, morphological normalization and word transformation to increase stemming performance. The stemmer is tested intrinsically using Paice's method and extrinsically on a basic tf-idf based IR system and an elementary news topic classifier using Multinomial Naive Bayes Classifier. The difference in performance of these systems with and without using the stemmer is analysed.

## I. Introduction

Stemming refers to the reduction of a given word into its stem which need not be the morphological root of the word. This is done to reduce the inflection of any particular word into a base form. For example: cats is the inflected form of cat and stemming strips the plurality suffix -s from cats to give cat.

Various NLP applications use stemming as a preprocessing step, for example: POS Tagging, Machine Translation, Document Clustering etc but arguably the most important role of word stemming is in Information Retrieval (IR). IR is an immensely common and important application of Natural Language Processing. It essentially refers to the retrieval of a particular document from a collection of documents.

There are two major problems while stemming: over-stemming and under-stemming. Over-stemming is when two separate inflected words are reduced to a same word stem. This is a false-positive in IR, since it leads the IR system to fetch documents which might not contain the search query. Similarly, under-stemming is when two same inflections of a word are not reduced to the same word stem. This is false-negative. It leads to an IR system not finding documents having a related word inflection.

Stemming is mostly done in three ways:
- Rule Based Stemming
- Statistical Stemming
- Hybrid Stemming

Rule based stemming approaches generally refer to affix stripping where a list of affixes are maintained and are stripped to stem a word. Similarly, statistical stemming refers to the usage of statistical models like HMMs and n-grams to stem a word. Hybrid stemming tends to combine aspects of both rule based stemming and statistical stemming in hopes of improving stemming performance. The focus of this work is on rule based method.

## II. Related Works

Stemming is not an unfamiliar topic. Including the renowned Porter stemmer, many works exist for stemming words in English. In Nepali, however, there are only a few works. Bal et al. wrote a morphological analyzer and stemmer for Nepali language [1]. Sitaula proposed a hybrid nepali stemming algorithm which uses affix stripping in conjunction with a string similarity function and reports a recall rate of 72.1% on 1200 words [2]. He has taken into consideration a total of 150 suffixes and around 35 prefixes. Paul et al. describes an affix removal stemming algorithm for Nepali text. Their work has a database of 120 suffixes and 25 prefixes and a root lexicon of over 1000 words and reports an overall accuracy of 90.48% [3]. Shrestha et al. classifies suffixes into three categories and stem them according to different criterias [4]. They take into account 128 suffix rules and report an accuracy of 88.78% on 5000 words.

There are also some works in languages which are morphologically similar to Nepali. A hindi stemmer was devised by Ramanathan et al. [5] where they first use a transliteration scheme to transliterate Devanagari to English. They have maintained a suffix list which is used to strip the word by using the process of longest match. Upon testing the algorithm in 35977 words, 4.6% words were found to be under-stemmed while 13.8% were found to be over-stemmed. An Urdu stemmer is also written by Kansal et al. [6] which uses the rule based approach to stem Urdu words. They report 85.14% accuracy on more than 20,000 words.

## III. CHALLENGES

The fact that Nepali is an inherently complex language makes it inaccessible to many analysis. Various derivational and inflectional techniques exist in Nepali grammar which creates plethora of frequently used words in everyday life. For instance, inflection alone is categorized as being of ten types. These inflections can alter a word's structure based on gender, cardinality, respect, tense and its aspects. Moreover, inflections are also based on moods, voice, causality and negation [7]. This makes it non-trivial to devise a proper stemming algorithm for Nepali language.

There is also a need to identify whether a linguistic entity attached at the end of the word is a suffix attaching itself to a base word or is actually a part of the word itself. For instance, in the word काले the entity ले is actually the part of the word itself whereas in the word कालेले the rightmost ले is a post-positional suffix. It is imperative to accurately identify when and when not to strip a given suffix because unnecessary stripping leads to over-stemming.

Another challenge in suffix stripping is the difference in writing. For example, both of the word form साङ्केतिक and साङ्केतीक are used interchangeably in informal writing. Unless an assumption about strictness of the grammar rules, there is a need to include both of the suffixes ि क and ी क. Not only that, several suffixes can be joined together as in उनीहरुको which contains two postpositions (हरु and को) compounded together. To deal with these scenarios, there is a need to repeatedly apply the stripping rules. However, this increases the chances of over-stemming.

## IV. METHODOLOGY

### A. Morphological Normalization

Among the vowels present in Nepali language, the vowel pairs <इ, ई> and <उ, ऊ> in both their dependant and in dependant forms are often confused while writing. Same is the case with some of the consonant groups like <व, ब>. To make the stemmer more robust to these common grammatical errors, a morphological normalization scheme was introduced where the often confused vowels and consonants are normalized into a single entity. Concretely, all occurrence of the vowel ई are replaced with इ and so on while stemming the words. A more detailed normalization scheme is outlined below.

### B. Prefix Stripping

Though there are many prefixes in Nepali, they have not been stripped as a part of this work. This is mainly because the prefixes derive a new word from a root instead of inflecting it. For instance, the words like उपकार, प्रकार, अधिकार, परिकार etc all are words derived from the application of the prefixes उप, प्र, अधि, परि respectively to the same root कार. All of these words are actually totally unrelated to each other so stripping prefixes would mean that they would overstem.

TABLE I
MORPHOLOGICAL NORMALIZATION RULES

| Vowel / Consonant | Normalized To |
|---|---|
| इ | ई |
| ी | ि |
| ऊ | उ |
| ू | ु |
| व | ब |
| श | स |
| ष | स |
| ँ | Nil (all occurances removed) |

An exception to this rule is the negativity prefix न. It usually occurs before verbs and negates their sense. For example, the verb जानु (to go) can be inflected as नजानु (to not go) by the application of this prefix. This work only considers this single prefix for stripping.

### C. Suffix Stripping

The suffixes in Nepali language have been classified into two classes in this work:

- Type I suffix
- Type II suffix

Type I suffixes mainly consists of post-positions and other agglutinative suffixes. Some example of these suffixes are: मा, बाट, ले, लाई, द्वारा, लागि, निम्ति etc. There are 85 type I suffixes identified in this work.

Type II suffixes, on the other hand, primarily consist of case markers and other bound suffixes. Some of the suffixes also occur in both free and bound form, for example ेका and एका are linguistically the same but differ in that the former has the dependant vowel े and the second has the independent vowel ए. Some examples of type II suffixes are: छे, ने, छ्यौ, एको, इक etc. A total of 161 of these suffixes were identified.

*1) Stripping type I suffix:* Stripping these suffixes is a non-trivial process. This can be attributed to two major facts:

To begin with, identification of these suffix is challenging. As was discussed earlier, some of these suffixes occur as a part of word itself. For instance, the word नेहरु is the name of a reputed Indian politician and not the suffix हरु attached to the root ने. There are many more examples of such exception words. Before stripping type I suffixes, an extensive exception word list has to be created and checked against to prevent over-stemming. A total of 181 of these exceptions words were identified by manually eyeballing a corpus derived from various online Nepali news sites. The corpus is described in section V-A.

Another challenge in stripping type I suffix is that these suffixes can be chained together i.e. the word उनीहरुलाई is a word created by chaining two different type I suffixes i.e. हरु and लाई. This requires repetitive stripping of the suffixes while checking the intermediate results against the exception word list.

*2) Stripping type II suffixes:* Stemming these suffixes is particularly tricky due to the inherent structure of Nepali Morphology. For example, consider the suffix इक. It is known to change the morphology of nouns in the following way:

सङ्गीत + इक = साङ्गीतिक

समाज + इक = सामाजिक

I.e. change of the dependent vowels (अ to आ ) at the start of the word.

To take these factors into consideration, we introduce a word transformation rule. In simple terms, if the word contains the इक prefix, the dependant vowel at the start of the word is changed accordingly. The vowel आ becomes अ, vowel औ becomes उ and the vowel ऐ becomes इ . Using this transformative rule, the word नैतिक would be transformed to the word नितिक . It is important to observe that this map does not map a word to its stem, rather only to an intermediate word, which will be then further processed to produce the correct stem. The intermediate word might not be grammatically correct one. The rationale being that the word नितिक and the word नीति would conflate to the same once they are morphologically normalized and then stemmed.

The stemming algorithm in itself is quite simple. In fact, after taking into account the variations in word morphology by addition of suffixes, the rest of the process is the repeated stripping of the suffixes in a longest suffix first approach. This stripping is done until further stripping is not possible. In the event that any particular stripping rule decreases the word size to below a set threshold, that rule is discarded. This is done to prevent over-stemming of the word. The threshold value for this project was taken to be 2 by observing the error rates as per the testing method described in section V-B.

## V. Performance Evaluation

### A. Data

To test the stemming rules and evaluate the over/under stemming errors, a corpus was constructed. This corpus was derived from various online news portals such as Setopati, Nagariknews, eKantipur etc. The corpus contained articles from various different areas including news, sports, politics, literature etc. Corpus contained a total of 4387 news articles with the total word count of 1181343 and total unique word count of 118056. Each news article, on average, contained 269 total words and 181 unique words.

### B. Intrinsic Evaluation - Paice's Method

Paice method [8] for evaluation of stemmers is based on under-stemming and over-stemming errors. In this method, a concept group is first defined where multiple word inflations of a single word-concept are grouped together. Similarly, a stemmer group is defined where words that produce same stem are grouped together. Using these two word groups, four performance indices are calculated and subsequent calculation of over-stemming index ($OI$) and under-stemming index ($UI$) is done. These indices and the method to calculate them are defined in [8].

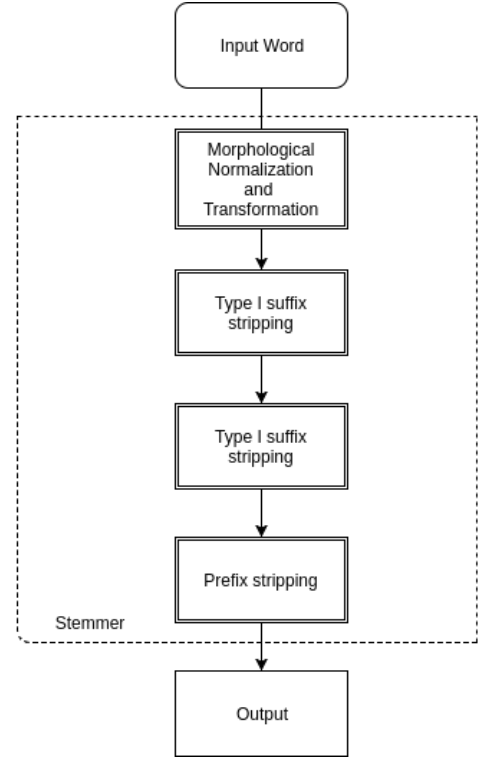

Fig. 1. Block diagram of the stemmer.

For evaluating the stemmer according to Paice method, 497 concept groups were defined. Each concept groups contained at least two related words with the maximum being thirty-nine words. A total of 1813 words constituted the concept groups. Some examples of the groups are as follows:

- तपाईँ, तपाई, तपाईँको, तपाईहरू, तपाईँले, तपाईको, तपाईले, तपाईंहरु
- हुनुपर्ने, हुनु, हुनुपर्छ, हुनुहुन्छ, हुनुहुन्थ्यो
- मानिस, मानिसको, मानिसहरू, मानिसलाई, मानिसले, मानिसमा, मानिसहरुको, मानिसहरुले

These words were derived from the top 10,000 most frequent words occurring in the corpus described in section V-A. The results obtained after running Paice method of evaluation on the stemmer using these concept groups are shown in table II.

Using these indices, the $OI$ was found to be 0.2% and the $UI$ was found to be 5.27%. This shows that the stemmer has high understemming error in contrast to over-stemming error implying that the stemmer is a *light stemmer* i.e. it has a tendency to not strip suffixes aggressively.

TABLE II
Paice Method Results

| Metric | Value |
|---|---|
| Global Desired Merge Total (GDMT) | 8274 |
| Global Unachieved Merge Total (GUMT) | 436 |
| Global Desired Non-Merge Total (GDNT) | 2742411 |
| Global Wrongly Merged Total (GWMT) | 4729 |

*C. Extrinsic Evaluation*

A most accurate and pragmatic test for any Stemmer is to actually implement a NLP application based on that Stemmer and then check for the performance of that application. For the purpose of this thesis, two different applications were designed. One of them being a crude IR system, which was developed using the Stemmer and then tested on a prepared dataset upon a subset of the corpus described in V-A. Another application was an elementary News Topic Classifier for seven different news topics.

*1) Information Retrieval Test:* Modern IR systems employ various measures like query expansion (where a simple input query is reconstructed to multiple queries for getting a wider coverage) to sophisticated relevancy algorithm like pagerank. For the purpose of this thesis, however, only a simple IR system has been developed where both documents and queries are modeled using the bag of words model and the ranking is done by using tf-idf metric which has been shown to give good results for document retrieval [9].

For the purpose of this test, total 100 documents were sampled from the corpus in 4.1. Then, 14 queries were constructed for retrieval. These queries contained one to three words and were constructed manually using the gathered documents. Some of the queries are shown below:
- पोखरीमा विष
- साझा बस
- कतार राजदुत
- अखिल क्रान्तिकारी

Using the TF-IDF ranking scheme, two independent information retrieval experiment were carried out for each query. The first experiment was done without stemming the documents and queries while the second experiment was done on the stemmed document and queries. The topmost result i.e. the document with the highest relevance score for the given query for both experiments were taken and three native Nepalese human judges were asked to assess the relevance of the retrieved document on the scale of one to five; one being the least relevant while five being most. If the query failed to return any document in any experiment, the relevance was taken to be zero.

The difference in average relevance score of the retrieved document with stemming and without stemming was calculated for each query and the differences were averaged at the end. The average gain in the relevance was found to be 0.93 i.e. 18.6%. The results of the experiment are summarized in fig. 2.

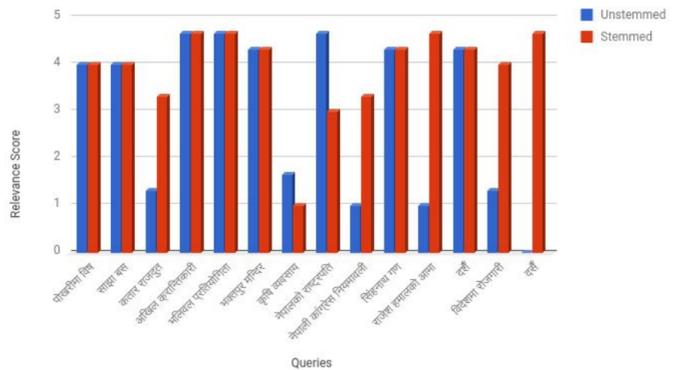

Fig. 2. Stemmed vs Non-Stemmed Relevance in IR experiment.

*2) News Topic Classification:* For the purpose of this particular application, a total of 1400 news articles belonging to seven categories like politics, economy, sports, literature, technology, global, and society were extracted from a Nepali news site nagariknews.com. Each topic contained 200 documents i.e. a uniform representation. A 70-30 split of training and test data was then done and a Multinomial Naive Bayes with Laplace smoothing was used for the subsequent classification.

A corpus wise stop word removal scheme was used i.e. terms appearing in more than half of the documents were removed and the tf-idf scheme [9] was used to construct a feature vector. The results for both stemmed and non-stemmed version of the classification is as follows:

TABLE III
Metrics for stemmed vs non-stemmed

| Scheme | Vocabulary Size | f1-score |
|---|---|---|
| Stemmed | 3217 | 0.79 |
| Non-stemmed | 5754 | 0.77 |

The F1 metric in Table III is a micro-averaged metric and since micro averaging in multiclass classification yields identical precision, recall, and f1; the precision and recall metrics are excluded from the table. The table clearly shows that in addition to significantly reducing the vocabulary size of the feature vector, stemmed classification also clearly outperforms the non-stemmed classification in terms of F1 score.